\pdfoutput=1

\documentclass[11pt]{article}

\usepackage[final]{acl}

\usepackage{times}
\usepackage{latexsym}
\usepackage{amsmath}
\usepackage{booktabs}
\usepackage{multirow}
\usepackage{multicol}
\usepackage{color}
\usepackage{float}

\usepackage{amssymb}
\usepackage{array}
\usepackage{graphicx}
\usepackage[T1]{fontenc}

\usepackage[utf8]{inputenc}

\usepackage{microtype}

\usepackage{inconsolata}

\usepackage{graphicx}
\usepackage{cleveref}
\title{DraDDP: A Multimodal Multi-Party Dialogue Discourse Parsing Dataset}

\author{Shannan Liu, Peifeng Li\thanks{ \ \ Corresponding author }, Yaxin Fan, Qiaoming Zhu \\
School of Computer Science and Technology, Soochow University, Suzhou, China \\
\texttt{20234027002@stu.suda.edu.cn} \\
\texttt{\{pfli, qmzhu\}@suda.edu.cn, yxfansuda@stu.suda.edu.cn}}

\begin{document}
\maketitle
\begin{abstract}
    Multi-party dialogue discourse parsing aims to identify dependency structures and relation types between utterances in conversations. Previous studies are mostly limited to textual modality or two-party dialogue, failing to meet the multimodal and multi-party settings. In this paper, we construct the first publicly available English multimodal dataset DraDDP for multi-party dialogue discourse parsing, based on American TV dramas. DraDDP contains 495 dialogue segments with 6,374 utterances and 9.1 hours of parallel video content, covering rich multi-party interaction scenarios. Moreover, we establish comprehensive benchmarks by evaluating this task on DraDDP and conducting in-depth analysis on the impact of different modalities. Experimental results demonstrate the value of multimodal information in capturing dialogue structures and relation types. We will publicly release the dataset, annotation guidelines, and code to promote future research in multimodal dialogue understanding.\footnote{\url{https://github.com/DraDDP}}
\end{abstract}

\begin{figure*}
  \includegraphics[width=1\textwidth]{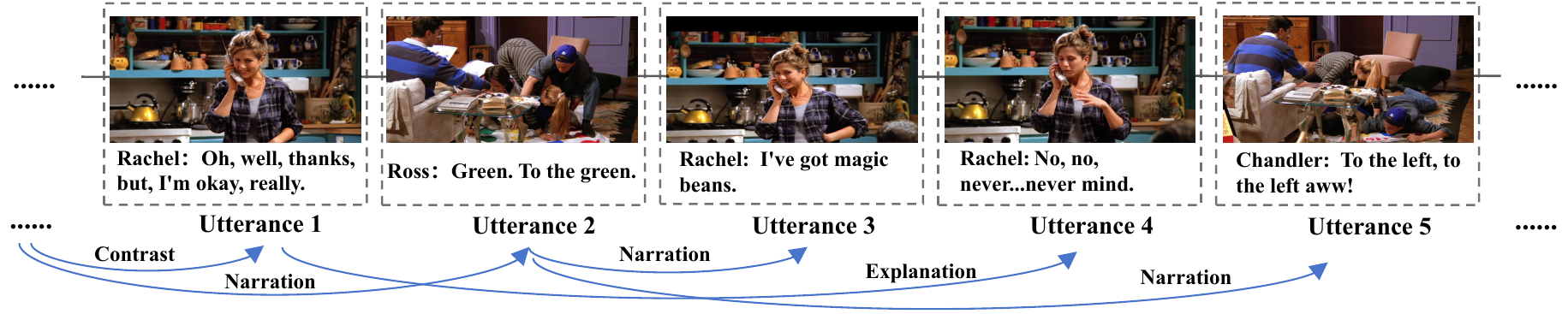}
\vspace{-0.6cm}  
  \caption{An example of multimodal dialogue discourse parsing.}
  \label{fig:teaser}
  \vspace{-0.1cm}  
\end{figure*}

\begin{table*}
\centering
\small
\resizebox{\textwidth}{!}{%
\begin{tabular}{lcccccc} 
\toprule
\textbf{Modalities} & \textbf{Dataset} & \textbf{\#Dialogues} & \textbf{\#Utterances} & \textbf{Data Source} & \textbf{Language} & \textbf{Participants} \\
\midrule
\textbf{T} & STAC & 1.1K & 10.7K & Online game & English & Multi-party \\
\textbf{T} & Molweni & 10K & 88.3K & Online forums & English & Multi-party \\
\textbf{T} & DialogueDSA & 705 & 24.5K & TV drama & Chinese & Multi-party \\
\textbf{T} & MSDC & 541 & 22.6K & Online game & English & Two-party \\
\textbf{T+I} & JDDC 2.1 & 246K & 3.46M & E-commerce platform & Chinese & Two-party \\
\textbf{T+V+A} & MODDP & 864 & 18K & TV drama & Chinese & Two-party \\
\textbf{T+V+A} & DraDDP & 495 & 6.4K & TV drama & English & Multi-party \\
\bottomrule
\end{tabular}%
}
\vspace{-0.1cm}  
\caption{Comparison with existing dialogue discourse parsing datasets (T/V/A/I: Text/Video/Audio/Image).}
\label{tab:dataset_comparison}
\vspace{-0.2cm}  
\end{table*}

\section{Introduction}
Multi-party dialogue discourse parsing aims to identify dependency structures and semantic relation types (e.g., \emph{Comment}, \emph{Background}, and \emph{Alternation}) between utterances in multi-party conversations. As shown in Figure~\ref{fig:teaser}, this example contains 5 utterances, where the arcs represent dependency structures between utterances, and the labels on the arcs indicate the types of discourse relations. This task is of significant value for downstream applications such as meeting summarization \cite{1,1.1,1.2}, dialogue generation \cite{2,2.1}, and emotion recognition \cite{3,3.1}.

Previous research on multi-party dialogue discourse parsing focused on two public datasets: STAC \cite{stac} and Molweni \cite{molweni}. However, these datasets only consider the textual modality, ignoring the complexity and richness of multimodal interactions in real-world scenarios \cite{7.1,7}. As shown in Figure~\ref{fig:teaser}, when relying solely on textual modality, it is difficult to understand why \emph{Ross} would suddenly mention \emph{``green''} (a seemingly unrelated response) after \emph{Rachel} expresses personal emotions. When audio-visual modalities are introduced, we observe that \emph{Rachel} is in a private phone call scenario, while \emph{Ross} is engaged in gaming interaction with friends nearby, with both being in parallel and independent dialogue contexts. This indicates that multimodal information not only supplements scene details not covered by text, but also plays an irreplaceable role in identifying dependency structures in multi-party dialogues and ensuring contextual semantic coherence.

Currently, there are only two available datasets for multimodal dialogue discourse parsing: JDDC 2.1 \cite{JDDC} and MODDP \cite{MODDP}. Both datasets focus on two-party dialogues and only support Chinese, failing to meet the needs of multi-party dialogue and cross-lingual research.
Compared to two-party dialogues, multi-party dialogues involve multiple participants and have more complex structures. Therefore, understanding the discourse structure of multi-party dialogues is more valuable and challenging.

In this paper, we construct the first English multimodal dataset DraDDP for the task of multi-party dialogue discourse parsing. DraDDP is annotated based on the classic American TV dramas (e.g., \emph{Friends}), covering rich multi-party interaction scenarios and emotional expression patterns. To the best of our knowledge, DraDDP is the first publicly available English multimodal multi-party dialogue discourse parsing dataset, providing a novel research benchmark for this field. The main contributions of this paper include:

1) We construct the first English multimodal dataset DraDDP for multi-party dialogue discourse parsing, containing 495 dialogue segments, 6,374 utterances, and 9.1 hours of parallel video content, providing rich resources for multimodal dialogue understanding research.

2) We establish comprehensive benchmarks by evaluating multiple dialogue discourse parsing models on DraDDP, and conducting systematic analysis to reveal the impact of multimodal information on parsing performance.

\section{Related Work}  
\textbf{Datasets} Table~\ref{tab:dataset_comparison} presents the core attributes of available datasets on dialogue discourse parsing. The textual datasets include STAC \cite{stac}, Molweni \cite{molweni}, DialogueDSA \cite{DialogueDSA}, and MSDC \cite{msdc}. Notably, while MSDC incorporates textually described non-verbal actions (e.g., ``picks up blue (-1,1,1)'') as discrete symbolic elementary discourse unit (EDU) nodes, real-world non-verbal signals (e.g., facial expressions and intonation) are continuous and temporally synchronized, exhibiting higher semantic complexity.

There are only two public datasets on multimodal discourse parsing: JDDC 2.1 \cite{JDDC} and MODDP \cite{MODDP}. Although JDDC 2.1 introduces image modality, the image content is relatively scarce and limited to specific domains. MODDP, sourced from TV drama dialogue scenarios, achieves significant improvements in modality completeness and scenario authenticity. However, both multimodal datasets only cover two-party dialogues and exclusively support Chinese, making it difficult to meet the urgent needs for multi-party dialogue and cross-lingual research.

\noindent \textbf{Textual Methods} Current mainstream research on dialogue discourse parsing primarily leverages pre-trained language models, which enhance parsing performance through strategies such as modeling key dialogue elements \cite{5.3,5.2}, injecting external information \cite{4,4.1}, or joint learning \cite{4.5,4.3}. With the rapid development of Large Language Models (LLMs), \citet{5}  and \citet{5.1} found that ChatGPT performed poorly on dialogue discourse parsing. \citeauthor{LLaMIPa} (\citeyear{LLaMIPa}) proposed LLaMIPa (LLaMA Incremental Parser), which achieved incremental prediction based on historical discourse structures through fine-tuning LLaMA3. Besides, \citeauthor{DDPE} (\citeyear{DDPE}) and \citeauthor{6} (\citeyear{6}) improved LLMs through explanatory prompts and dialogue clarification. However, these advances remain limited to text-only scenarios.

\noindent \textbf{Multimodal Methods} Only MODDP \cite{MODDP} provides a basic multimodal benchmark, employing cross-modal attention  to fuse multimodal features. However, it focuses solely on two-party Chinese dialogues and does not explore multimodal large language models (MLLMs). We address this gap by constructing the first English multimodal dataset for the task of multi-party dialogue discourse parsing and benchmarking both traditional and MLLM approaches.

\section{Data Construction}
\subsection{Data Preparation}
DraDDP uses the first season of the American TV series \textit{Friends} (1994) as its data source, covering all 24 episodes. This choice offers two main advantages: 1) the dialogue participants include 6 core protagonists and over 20 supporting characters, generating rich multimodal interaction information such as body language, facial expressions, and intonation; 2) the dialogue scenarios are diverse, covering homes, cafes, and workplaces, with topics spanning multiple dimensions including emotions, daily life, humor, and negotiation, providing representative multi-party interaction patterns for discourse parsing research.

For EDU segmentation, we adopted each official subtitle line as a basic discourse unit based on three considerations: subtitle lines are professionally produced with moderate length; segmentation follows speaker turns and semantic boundaries without crossing scene transitions; and precise timestamps enable accurate alignment of text, video frames, and audio segments.

\subsection{Annotation Guidelines}
The dialogue datasets shown in Table~\ref{tab:dataset_comparison} are all constructed based on Segmented Discourse Representation Theory (SDRT) \cite{SDRT}. SDRT employs directed graph structures to represent dependency relations between discourse units, which can effectively capture complex interaction patterns and dynamic contextual changes in dialogues.
Furthermore, we utilized the 16 relation labels from the STAC \cite{stac} system (detailed in Appendix~\ref{sec:appendix1}) to distinguish different types of discourse relations. This labeling system provides comprehensive definitions and rich annotation examples, offering reliable guarantees for annotation quality assurance.

\subsection{Annotation Quality Control}
\label{subsec:3.3}
To ensure the dataset annotation quality meets academic standards, we designed a rigorous four-stage quality control system. The entire annotation work was completed through collaboration among 2 doctoral students and 4 master's students, whose research fields are dialogue or discourse analysis. 

1) We introduced a pre-annotation mechanism to improve data annotation efficiency and provide reliable initial references for human annotation. Specifically, we employed LLaMA3\footnote{\url{https://huggingface.co/meta-llama/Meta-Llama-3-8B}} fine-tuned on the STAC dataset to perform preliminary discourse structure prediction on all data based on the text modality (\cref{subsec:pre-annotation} for details).

2) The annotators collectively watched corresponding video segments based on model predictions and collaboratively corrected discourse structures. 
The main goal was to establish unified annotation standards, during which 1/6 of the dataset was annotated collaboratively. We systematically compiled problems encountered during annotation and developed comprehensive annotation guidelines, which will be publicly released as supplementary materials alongside the dataset. 

3) Each episode's data was randomly assigned to two different annotators to independently complete full discourse structure annotation. When the results from both annotators were completely consistent, the annotation was directly adopted as the final result; when disagreements occurred, consensus was reached through collective discussion, and related issues were incorporated into the annotation guidelines for improvement. This stage completed annotation of 1/3 of the dataset. 

4) The remaining data was randomly assigned to two annotators for preliminary annotation, with disagreed portions adjudicated by a third annotator (a doctoral student).

We used Fleiss' Kappa coefficient \cite{9} to evaluate inter-annotator agreement. The Kappa value for discourse dependency structures was 0.91, showing high consistency, mainly attributed to most dependency structures occurring between adjacent utterances, making identification relatively straightforward. The Kappa value for relation types was 0.60, which, although exceeding the STAC corpus's 0.58, reflects the inherent complexity and challenges of distinguishing discourse relations through its relatively low consistency. We present several challenging annotation cases in Appendix~\ref{sec:appendix2} to further illustrate the difficulties and complexities encountered during the annotation process.

\begin{figure}
    \centering
    \includegraphics[width=0.9\linewidth]{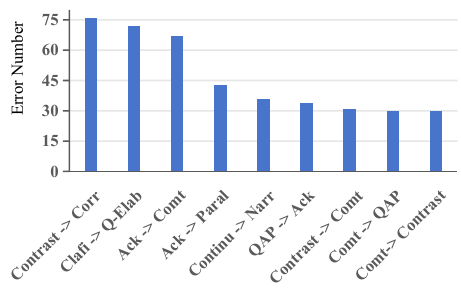} 
    \vspace{-0.4cm}
    \caption{Error statistics from pre-annotation where \emph{\{X $\to$ Y\}} indicates misclassifying relation X as Y and the relations are defined in Appendix~\ref{sec:appendix1}.}
    \label{fig:Pre-annotation analysis error statistics}
    \vspace{-0.2cm}
\end{figure}

\subsection{Pre-annotation Analysis on LLaMA3}
\label{subsec:pre-annotation}
To improve annotation efficiency and provide initial references for human annotators, we employed a pre-annotation mechanism. Specifically, we fine-tuned the LLaMA3 model on the STAC dataset to perform preliminary discourse structure prediction on DraDDP based on textual modality. The model takes the text sequence from the dialogue start to the current utterance $\{u_0, u_1, \ldots, u_i\}$ as input and outputs the dependency parent node and relation type for the current utterance $u_i$.

Compared with final human annotations, the model achieved an F1 score of 72.69\% on dependency structures and 41.31\% on relation types. It is important to emphasize that the pre-annotation results served only as a reference and were not decisive. Annotators primarily made judgments based on watching video clips and text content independently. During the pre-annotation correction phase, annotators made an average of 3.8 modifications to the dependency structure of each dialogue segment and an average of 7.4 adjustments to the relation types. To further verify that pre-annotation did not introduce systematic bias, we conducted a controlled experiment: selecting 50 dialogue samples, two groups of annotators independently completed the same annotation tasks, with one group correcting pre-annotations and the other annotating from scratch. The inter-group Kappa value reached 0.90 for dependency structure and 0.58 for relation type, almost consistent with the overall annotation reported in \cref{subsec:3.3}. Given that nearly 82\% of dependency relations occur between adjacent utterances, the pre-annotation model more easily captures short-distance dependencies, thereby significantly reducing annotators' workload on short-distance dependency structures.

As shown in Figure~\ref{fig:Pre-annotation analysis error statistics}, the model most frequently misclassifies \emph{Contrast} as \emph{Correction} (76 instances). Both involve responsive expressions to preceding content. However, \emph{Contrast} emphasizes different viewpoints or situations while \emph{Correction} explicitly indicates modifications of previous information. The model struggles to accurately capture these subtle semantic distinctions. Another significant error is misclassifying \emph{Clarification\_Question} as \emph{Q-Elab} (72 instances), further demonstrating the inadequacy of text-based models in handling semantically similar relation types.

\begin{figure}
    \centering
    \includegraphics[width=1\linewidth]{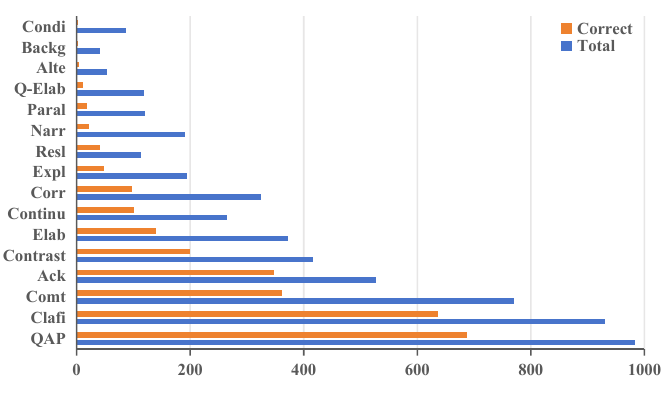} 
    \vspace{-0.8cm}
    \caption{Analysis of pre-annotation results, including total number and correct number of each relation type.}
    \label{fig:total number and correct number}
    \vspace{-0.2cm}
\end{figure}

Nevertheless, as shown in Figure~\ref{fig:total number and correct number}, the model performs well on structurally regular discourse relations (e.g., \emph{Question-Answer Pair (QAP)} and \emph{Clarification\_Question}). This indicates that LLMs can identify regular discourse patterns but still show deficiencies in distinguishing semantically similar categories, highlighting the importance of introducing multimodal information for improving dialogue discourse parsing performance.

In summary, pre-annotation can effectively reduce annotators' workload on short-distance dependency structures and regular discourse relations without compromising annotation quality.

\begin{figure*}
    \centering
    \includegraphics[width=1\linewidth]{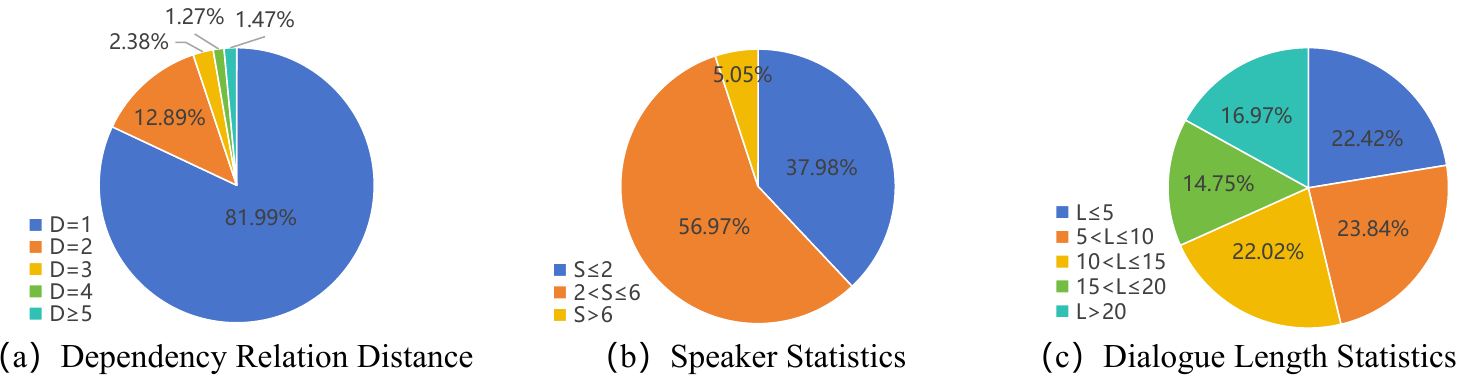}
    \caption{Data analysis of the DraDDP dataset.}
    \label{fig:Data Analysis}
\end{figure*}

\section{Dataset Analysis}
To gain deeper insights into the feature distribution and complexity of the DraDDP dataset, we conducted statistical analysis across four dimensions:  dependency distance, dialogue participant scale, dialogue length, and label distribution.

\subsection{Dependency Distance}
Dependency distance represents the positional span between the ``current utterance'' and its ``parent utterance'', serving as a key indicator of dialogue semantic coherence and structural complexity. As shown in Figure~\ref{fig:Data Analysis}(a), 81.99\% of dependencies have a distance of 1, meaning most dependencies occur between adjacent utterances. Compared to the dyadic dialogue dataset MODDP (where ``D = 1'' accounts for 92.0\%), DraDDP shows a lower proportion. This indicates that when the number of participants expands from ``dyadic'' to ``multi-party'', dialogue topics are more prone to branching and jumping, making the semantic structure of multi-party dialogues more ``non-local''.

\subsection{Speaker Statistics}
We analyze the speakers in each dialogue segment. As shown in Figure~\ref{fig:Data Analysis}(b), single/two-party dialogues account for 37.98\%, 3-6 person conversations account for 56.97\% (the highest proportion), and conversations with more than 6 participants account for 5.05\%. This aligns with the setting of \textit{Friends}, which centers on the daily interactions of 6 core protagonists, making 3-6 person conversations the predominant dialogue type. The increase in participants leads to models needing to traverse more dialogue history turns to capture authentic discourse dependency structures (see Appendix~\ref{sec:appendix3}), significantly increasing parsing difficulty.

\subsection{Dialogue Length Statistics}
Dialogue length determines the span of semantic context and serves as an important metric for evaluating models' long-text modeling capabilities. As shown in Figure~\ref{fig:Data Analysis}(c), the distribution across different dialogue length intervals is relatively uniform, with an average dialogue length of 12.88. This uniform distribution is beneficial for effective model training in different dialogue length scenarios.

\subsection{Label Distribution}
Appendix~\ref{sec:appendix1} presents the distribution of discourse relation types across DraDDP and existing datasets. The \emph{Question-Answer Pair} relation dominates in multi-party datasets (DraDDP and STAC) and accounts for 9.4\% in dyadic MODDP, indicating that question-answer exchanges serve as core communication patterns for introducing topics and exchanging information. Compared to STAC where the top two relations account for 41.8\%, DraDDP (34.6\%) and MODDP (32.1\%) show more balanced distributions. This difference stems from data sources: STAC's game scenario concentrates relations on question-answer and comment patterns, while DraDDP and MODDP from TV dramas encompass richer social interactions and emotional expressions. Additionally, multimodal cues (tone, facial expressions) help identify non-core relations that are difficult to understand through text alone (Details in \cref{subsec:Error Pattern Analysis} and \cref{subsec:Case Study}), further promoting label balance.

\section{Approach}
\subsection{Problem Definition}
Given a dialogue sequence $D = \{U_0, U_1, \ldots, U_n\}$, where each utterance $U_i$ contains the textual modality $U_i^t$, the audio modality $U_i^a$, and the video modality $U_i^v$, our goal is to predict the utterance dependency graph $G = \{(U_i, U_j : R_{ij}) \mid i < j, R_{ij} \in \mathcal{R}\}$, where $(U_i, U_j : R_{ij})$ represents a dependency arc from utterance $U_i$ to $U_j$ with the relation type $R_{ij}$, and $\mathcal{R}$ is the predefined set of relation types in Appendix~\ref{sec:appendix1}.

\begin{table*}[htbp]
\centering
\setlength{\tabcolsep}{11.6pt}
\begin{tabular}{llcccc}
\toprule
\multirow{2}{*}{Category} & \multirow{2}{*}{Model} & \multicolumn{2}{c}{DraDDP} & \multicolumn{2}{c}{MODDP} \\
\cmidrule(lr){3-4} \cmidrule(lr){5-6}
& & Link & Link\&Rel & Link & Link\&Rel \\
\midrule
\multirow{6}{*}{Unimodal} 
& RLTST & 76.94 & 40.10 & 90.76 & 41.41 \\
& BERTLine & 78.65 & 45.82 & 90.73 & 40.96 \\
& MODDP$_{\text{T}}$ & 83.31 & 45.06 & \textbf{92.46} & 43.66 \\
& LLaMIPa & 84.71 & 53.39 & 91.32 & 53.05 \\
& LLaMIPa$^\dagger$ & \textbf{85.03} & 54.58 & 91.55 & 53.68 \\
& LLaMIPa$^\ddagger$ (Qwen2.5) & 84.14 & 53.55 & 91.26 & 52.82 \\
\midrule
\multirow{4}{*}{Multimodal} 
& MODDP$_{\text{Multi}}$ & 83.65 & 46.06 & 90.90 & 48.05 \\
& LLaMIPa$^\ddagger$ (Qwen2.5-VL) & 84.15 & 53.15 & 91.33 & 51.11 \\
& LLaMIPa$^\ddagger$ (Qwen2-Audio) & 84.90 & \textbf{55.09} & 92.43 & \textbf{54.88} \\
& LLaMIPa$^\ddagger$ (Qwen2.5-Omni) & 84.55 & 53.34 & 91.46 & 52.27 \\
\bottomrule
\end{tabular}%

\caption{Experimental results (F1 scores) on DraDDP and MODDP test sets. $^\dagger$: Without concatenating dialogue structure history previously predicted by the model. $^\ddagger$: Built upon the LLaMIPa$^\dagger$ framework with the backbone model replaced by Qwen series (7B). Qwen2.5-VL: Text+Video; Qwen2-Audio: Text+Audio; Qwen2.5-Omni: Text+Video+Audio.}
\label{tab:experimental_results}
\vspace{-0.2cm}
\end{table*}

\begin{table}[htbp]
\centering
\resizebox{\linewidth}{!}{%
\begin{tabular}{lcccc}
\toprule
Category & Train & Dev & Test & Total \\
\midrule
Dialogues & 345 & 75 & 75 & 495 \\
Utterances & 4,447 & 962 & 965 & 6,374 \\
Avg. Utt/Dialog & 12.89 & 12.83 & 12.87 & 12.88 \\
Avg. Utt Length & 9.54 & 10.19 & 9.59 & 9.64 \\
\bottomrule
\end{tabular}
}
\caption{Data statistics of the DraDDP dataset.}
\label{tab:DraDDP_statistics}
\vspace{-0.2cm}
\end{table}

\subsection{Benchmarks}
We employed four advanced dialogue discourse parsing systems to validate DraDDP's effectiveness: \textbf{RLTST}~\cite{4.2}, integrates multi-task learning with reinforcement learning to address data sparsity; \textbf{BERTLine}~\cite{4.7}, fine-tunes BERT to encode EDU pairs and handle multi-parent structures; \textbf{MODDP}~\cite{MODDP} fuses text, visual, and audio features via cross-modal attention with context-aware modules; and \textbf{LLaMIPa}~\cite{LLaMIPa} performs incremental parsing by incorporating historical discourse structures during fine-tuning LLaMA3 (details in Appendix~\ref{sec:appendix3.1}).

Moreover, we introduced two variants of LLaMIPa: \textbf{LLaMIPa$^\dagger$} removes the historical structure concatenation mechanism from the original LLaMIPa. \textbf{LLaMIPa$^\ddagger$} adopts the LLaMIPa$^\dagger$ framework while replacing the backbone LLaMA3 (8B version) with Qwen-series (7B version).

\section{Experimentation}
\subsection{Experimental Setup}
\textbf{Data} As shown in Table~\ref{tab:DraDDP_statistics}, we randomly split DraDDP proportionally into training, development, and test sets, and evaluate the above benchmarks on both DraDDP and  MODDP (using the split in  \citet{MODDP}).

\noindent \textbf{Evaluation Metrics} Following previous research, we use micro F1 scores as evaluation metrics: \textbf{Link-F1} evaluates the accuracy of dependency edge identification; \textbf{Link\&Rel-F1} requires both edges and relation types to be correct.

\noindent \textbf{Implementation Details} All experiments were conducted on a server equipped with two NVIDIA RTX 4090D GPUs. All LLMs were fine-tuned with LoRA using the LLaMA-Factory framework\footnote{\url{https://github.com/hiyouga/LLaMA-Factory}}, setting the rank to 8 and the scaling parameter to 16. We employed the AdamW optimizer with a learning rate of $1 \times 10^{-4}$. The batch size per GPU was set to 1 with gradient accumulation steps of 8. All models were trained for 3 epochs using mixed precision training. For video processing, frames were sampled at 1 fps (up to 16 frames maximum). Audio signals were encoded after converting 16 kHz sampled audio into 80-channel Mel spectrograms. During training, checkpoints were saved every 500 steps, and we retained the model with the best Link\&Rel-F1 performance on the development set.

\begin{table*}[htbp]
\centering
\resizebox{\linewidth}{!}{%
\begin{tabular}{lcccccccc}
\toprule
\multirow{2}{*}{\textbf{Model}} & \multicolumn{2}{c}{\textbf{LLaMIPa$^\ddagger$ (Qwen2.5)}} & \multicolumn{2}{c}{\textbf{LLaMIPa$^\ddagger$ (Qwen2.5-VL)}} & \multicolumn{2}{c}{\textbf{ LLaMIPa$^\ddagger$ (Qwen2-Audio)}} & \multicolumn{2}{c}{\textbf{LLaMIPa$^\ddagger$ (Qwen2.5-Omni)}} \\
\cmidrule(lr){2-3} \cmidrule(lr){4-5} \cmidrule(lr){6-7} \cmidrule(lr){8-9}
& Link & Link\&Rel & Link & Link\&Rel & Link & Link\&Rel & Link & Link\&Rel \\
\midrule
$s\leq2$ & 88.87 & 53.20 & 88.28 & 55.28 & 88.8 & 54.09 & 89.17 & 55.05 \\
$2<s\leq6$ & 80.80 & 53.30 & 81.60 & 51.30 & 83.20 & 54.90 & 80.60 & 48.90 \\
$s>6$ & 79.85 & 55.42 & 81.77 & 51.58 & 87.54 & 61.19 & 87.54 & 59.27 \\
\bottomrule
\end{tabular}%
}
\vspace{-0.2cm}
\caption{Performance on DraDDP with different models and number of speakers.}
\label{tab:speaker_multimodal}
\vspace{-0.2cm}
\end{table*}

 \begin{figure*}
    \centering
    \includegraphics[width=0.83\linewidth]{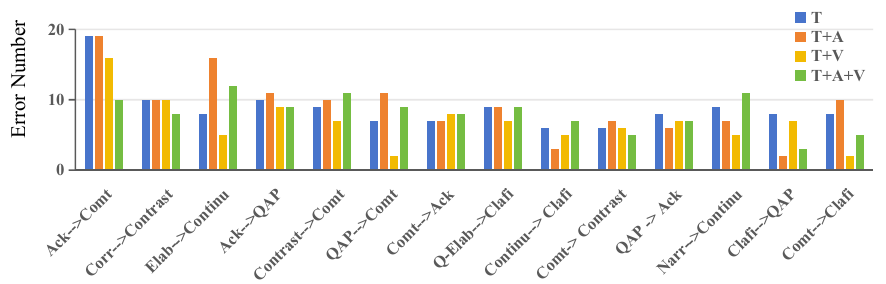}
    \vspace{-0.2cm}
    \caption{Error pattern statistics of the DraDDP test set under different modalities, where \emph{\{X $\to$ Y\}} indicates misclassifying relation X as Y.}
    \label{fig:Error mode statistics of DraDDP test set in different modes}
    \vspace{-0.1cm}
\end{figure*}

\subsection{Experimental Results}
\label{subsec:Experimental Results}
Table~\ref{tab:experimental_results} presents the results on the DraDDP and MODDP datasets. Compared to dyadic dialogues, multi-party dialogues exhibit higher complexity. On the DraDDP dataset, LLaMIPa$^\ddagger$ (Qwen2.5) achieves a Link-F1 of only 84.14\%, which is 7.12 lower than on the MODDP dataset, reflecting the inherent challenges posed by the more complex discourse interaction structures in multi-party dialogue scenarios.

Comparing incremental  LLaMIPa with non-incremental  LLaMIPa$^\dagger$, we observe that removing history concatenation improves Link\&Rel-F1 performance by 1.19, indicating that early prediction errors can mislead subsequent parsing decisions. Therefore, we adopt the LLaMIPa$^\dagger$ framework for subsequent experiments.

Introducing the audio modality can significantly improve performance. On the DraDDP dataset, LLaMIPa$^\ddagger$ (Qwen2-Audio) achieves a 1.54 improvement in Link\&Rel-F1 over LLaMIPa$^\ddagger$ (Qwen2.5), and a 2.06 improvement on the MODDP dataset. This demonstrates that audio cues (e.g., intonation, emotional tendencies, and voice intensity) can provide fine-grained relation discrimination information beyond text.

In contrast, the visual modality shows limited effectiveness. The reasons are: 1) the current video encoding approach (1 fps sampling) struggles to capture temporal variations in fine-grained facial expressions and body movements; 2) visual information in TV drama scenes contains substantial background noise unrelated to discourse relations. This reveals that in multimodal dialogue discourse parsing, different modalities contribute significantly differently, requiring targeted design of modality fusion strategies. We further conduct comprehensive ablation studies to quantify the contribution of each modality in Appendix~\ref{sec:appendix5.1}. Additionally, we evaluate the performance of the SOTA LLMs including GPT-4o and Claude in Appendix~\ref{sec:appendix5}.

\begin{figure*}
    \centering
    \includegraphics[width=0.83\linewidth]{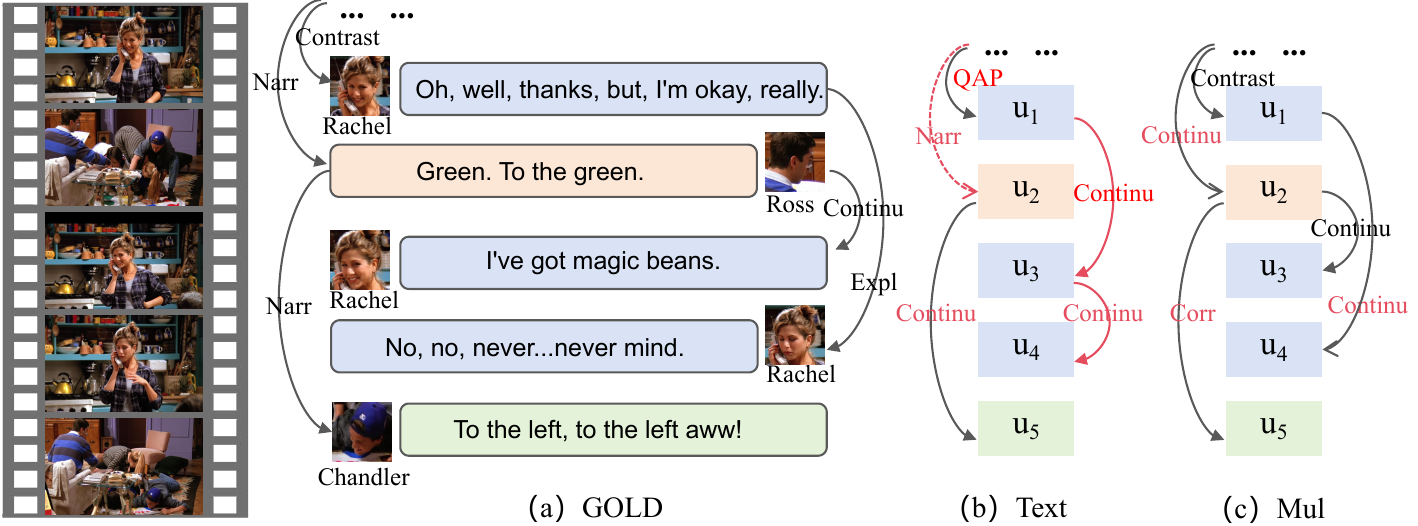} 
    \caption{A case study on unimodal (T) and multimodal (T+V+A) dialogue parsing using the DraDDP dataset.}
    \label{fig:example of dialogue discourse parsing}
\end{figure*}

Notably, we further analyze the performance of different modalities across varying speaker counts. As shown in Table~\ref{tab:speaker_multimodal}, we divide the DraDDP test set into three groups based on participant numbers ($s\leq2$, $2<s\leq6$, $s>6$). 

In the $s\leq2$ scenario, the video modality contributes significantly to relation type identification. LLaMIPa$^\ddagger$ (Qwen2.5-VL) outperforms the pure text model LLaMIPa$^\ddagger$ (Qwen2.5) by 2.08 of Link\&Rel-F1. This is because in two-party dialogues, visual cues such as facial expressions and eye contact are more focused and prominent, effectively assisting in relation type judgment. 

In the $2<s\leq6$ scenario, the audio modality performs better. LLaMIPa$^\ddagger$ (Qwen2-Audio) achieves improvements of 2.40 and 1.60 in Link-F1 and Link\&Rel-F1, respectively, compared to the pure text model. This indicates that as dialogue complexity increases, audio features such as intonation variations and emotional coloring begin to provide valuable discriminative information. However, the multimodal fusion model (LLaMIPa$^\ddagger$ (Qwen2.5-Omni)) shows a performance decline to 48.90\% in Link\&Rel-F1, suggesting that current multimodal LLMs still have significant deficiencies in effectively fusing audio and visual information. The modality fusion introduces noise interference, weakening the model's ability to identify both dependency structures and relation types.

In the complex multi-party dialogue scenario with $s>6$, the advantage of audio modality becomes more pronounced, achieving improvements of 7.69 and 5.77 over the pure text model. This indicates that as the number of dialogue participants increases, the audio modality can more effectively help the model identify discourse dependency structure and semantic relation types in complex multi-party interactions by capturing features such as speakers' tone variations and emotional intensity.

\subsection{Analysis on Relation and Modality}
\label{subsec:Error Pattern Analysis}
To explore the impact of different modalities on discourse relation types, we analyzed typical error patterns under different modality combinations using the LLaMIPa$^\ddagger$ (Qwen2.5-Omni) model, as shown in Figure~\ref{fig:Error mode statistics of DraDDP test set in different modes}.

Audio demonstrates significant advantages in distinguishing emotion-oriented and question-oriented relations. For example, \emph{\{Comt $\to$ Clafi\}} errors decrease by 71.4\%, as \emph{Clafi} typically carries confused questioning intonation while \emph{Comt} exhibits clear emotional tendencies (e.g., sarcasm, praise, criticism). Similarly, \emph{\{QAP $\to$ Comt\}} errors decrease by 75\%, since answers to questions usually maintain stable intonation whereas comments carry richer emotional coloring.

Visual cues play an important role in identifying certain interactive discourse relations. For example, \emph{\{Continu $\to$ Clafi\}} errors decrease by 50\% and \emph{\{Clafi $\to$ QAP\}} by 75\%. The visual modality provides crucial cues through facial expressions, gestures, and body language. For example, sustained eye contact and unchanged gaze help identify \emph{Continu}, while confused expressions and questioning gestures help distinguish \emph{Clafi} from \emph{QAP}. However, it also increases certain errors such as \emph{\{Elab $\to$ Continu\}} and \emph{\{QAP $\to$ Comt\}}, as visual cues (e.g., sustained eye contact) may sometimes be misinterpreted as topic continuation signals, and expressions during answers may be mistaken for emotionally evaluative behaviors.

When simultaneously introducing audio and visual information, \emph{\{Ack $\to$ Comt\}} decrease by 47.4\%, \emph{\{Clafi $\to$ QAP\}} by 62.5\%, and \emph{\{Comt $\to$ Clafi\}} by 35\%. These improvements demonstrate the complementary nature of audio-visual fusion. However, certain error types increase, such as \emph{\{Elab $\to$ Continu\}} by 50\% and \emph{\{Contrast $\to$ Comt\}} by 22.2\%, suggesting that multimodal fusion may cause information interference for specific relation types. For instance, synchronized body movements and speech rhythms may lead to misidentifying elaborative content as simple continuation, while contrastive tones combined with rich facial expressions may be confused with evaluative comments.

Overall, multimodal information provides valuable semantic cues for emotion-related and interaction-related discourse relations, though not universally beneficial. Future research should focus on refined fusion strategies to maximize modality advantages while minimizing information conflicts.

\subsection{Case Study}
\label{subsec:Case Study}
To demonstrate the role of multimodal information in dialogue discourse parsing, we analyzed a typical case from DraDDP. As shown in Figure~\ref{fig:example of dialogue discourse parsing}, we compare LLaMIPa$^\ddagger$ (Qwen2.5-Omni) performance under unimodal and multimodal settings.

In this scenario, the unimodal model fails to identify the sudden topic shift and cannot understand why $u_{2}$ would suddenly mention content related to ``green'' after $u_{1}$ expresses personal emotions. With the help of multimodal information ($u_{3}$ shows Rachel looking at Ross while speaking), the model identifies the dependency structures of $u_{2}$-$u_{3}$ and $u_{1}$-$u_{4}$, where the mentioned \emph{green} and \emph{magic beans} indicate a gaming interaction scenario, while $u_{1}$ and $u_{4}$ are in a private phone call scenario. This result demonstrates that in multi-party dialogue discourse parsing, effectively integrating multimodal information such as audio and video that contains ``interactive intentions'' and ``scene associations'' can enable more precise parsing of complex dialogue semantic structures.

\section{Conclusion}
This paper fills the research gap in multimodal multi-party English dialogue discourse parsing by constructing the DraDDP dataset. We establish comprehensive benchmarks by evaluating multiple state-of-the-art models, including traditional discourse parsing systems and LLMs. Experiments and analysis validate the importance of multimodal information in understanding dialogue semantics and relation types, providing data support and technical reference for subsequent research in this field.

\section{Limitations}
We identify two main limitations of our work. First, despite substantial annotation efforts, the DraDDP dataset remains relatively small due to the inherent complexity of multimodal multi-party dialogue discourse parsing and intensive annotation requirements. The annotation process requires simultaneous consideration of textual content, audio-visual cues, speaker interactions, and discourse structures, which is significantly more time-consuming than traditional text-only annotation tasks. Second, the dataset is sourced from a specific TV series, which may carry domain bias from sitcom-specific dialogue patterns, humor styles, and scripted dialogue dynamics. Future work should focus on expanding dataset scale and diversifying data sources to cover more spontaneous, cross-cultural, and contemporary dialogue scenarios.



\bibliography{custom}

\begin{thebibliography}{30}
\providecommand{\natexlab}[1]{#1}

\bibitem[{Asher et~al.(2016)Asher, Hunter, Morey, Benamara, and Afantenos}]{stac}
Nicholas Asher, Julie Hunter, Mathieu Morey, Farah Benamara, and Stergos Afantenos. 2016.
\newblock Discourse structure and dialogue acts in multiparty dialogue: the stac corpus.
\newblock In \emph{Proceedings of the 10th International Conference on Language Resources and Evaluation}, pages 2721--2727.

\bibitem[{Asher and Lascarides(2003)}]{SDRT}
Nicholas Asher and Alex Lascarides. 2003.
\newblock \emph{Logics of conversation}.
\newblock Cambridge University Press.

\bibitem[{Bennis et~al.(2023)Bennis, Hunter, and Asher}]{4.7}
Zineb Bennis, Julie Hunter, and Nicholas Asher. 2023.
\newblock A simple but effective model for attachment in discourse parsing with multi-task learning for relation labeling.
\newblock In \emph{Proceedings of the 17th Conference of the European Chapter of the Association for Computational Linguistics}, pages 3412--3417.

\bibitem[{Chan et~al.(2023)Chan, Cheng, Wang, Jiang, Fang, Liu, and Song}]{5}
Chunkit Chan, Jiayang Cheng, Weiqi Wang, Yuxin Jiang, Tianqing Fang, Xin Liu, and Yangqiu Song. 2023.
\newblock Chatgpt evaluation on sentence level relations: A focus on temporal, causal, and discourse relations.
\newblock \emph{arXiv preprint arXiv:2304.14827}.

\bibitem[{Fan et~al.(2023)Fan, Jiang, Li, Kong, and Zhu}]{4.2}
Yaxin Fan, Feng Jiang, Peifeng Li, Fang Kong, and Qiaoming Zhu. 2023.
\newblock Improving dialogue discourse parsing via reply-to structures of addressee recognition.
\newblock In \emph{Proceedings of the 2023 Conference on Empirical Methods in Natural Language Processing}, pages 8484--8495.

\bibitem[{Fan et~al.(2024{\natexlab{a}})Fan, Jiang, Li, and Li}]{5.1}
Yaxin Fan, Feng Jiang, Peifeng Li, and Haizhou Li. 2024{\natexlab{a}}.
\newblock Uncovering the potential of chatgpt for discourse analysis in dialogue: An empirical study.
\newblock In \emph{Proceedings of the 2024 Joint International Conference on Computational Linguistics}, pages 16998--17010.

\bibitem[{Fan et~al.(2025{\natexlab{a}})Fan, Li, Kong, and Zhu}]{4.3}
Yaxin Fan, Peifeng Li, Fang Kong, and Qiaoming Zhu. 2025{\natexlab{a}}.
\newblock Enhancing multiparty dialog discourse parsing with dynamic task-adaptive graph transformer and difficulty-aware task scheduling.
\newblock \emph{IEEE Transactions on Neural Networks and Learning Systems}, 36(9):16492--16506.

\bibitem[{Fan et~al.(2024{\natexlab{b}})Fan, Li, and Zhu}]{2}
Yaxin Fan, Peifeng Li, and Qiaoming Zhu. 2024{\natexlab{b}}.
\newblock Improving multi-party dialogue generation via topic and rhetorical coherence.
\newblock In \emph{Proceedings of the 2024 Conference on Empirical Methods in Natural Language Processing}, pages 3240--3253.

\bibitem[{Fan et~al.(2025{\natexlab{b}})Fan, Li, and Zhu}]{6}
Yaxin Fan, Peifeng Li, and Qiaoming Zhu. 2025{\natexlab{b}}.
\newblock Improving dialogue discourse parsing through discourse-aware utterance clarification.
\newblock \emph{arXiv preprint arXiv:2506.15081}.

\bibitem[{Feng et~al.(2021)Feng, Feng, Qin, and Geng}]{1}
Xiachong Feng, Xiaocheng Feng, Bing Qin, and Xinwei Geng. 2021.
\newblock Dialogue discourse-aware graph model and data augmentation for meeting summarization.
\newblock In \emph{Proceedings of the Thirtieth International Joint Conference on Artificial Intelligence}, pages 3808--3814.

\bibitem[{Fleiss(1971)}]{9}
Joseph~L Fleiss. 1971.
\newblock Measuring nominal scale agreement among many raters.
\newblock \emph{Psychological bulletin}, 76(5):378.

\bibitem[{Gao et~al.(2023)Gao, Cheng, Li, Chen, Li, Zhao, and Yan}]{1.1}
Shen Gao, Xin Cheng, Mingzhe Li, Xiuying Chen, Jinpeng Li, Dongyan Zhao, and Rui Yan. 2023.
\newblock Dialogue summarization with static-dynamic structure fusion graph.
\newblock In \emph{Proceedings of the 61st Annual Meeting of the Association for Computational Linguistics}, pages 13858--13873.

\bibitem[{Gong et~al.(2024)Gong, Kong, Zhao, Li, and Fu}]{MODDP}
Chen Gong, DeXin Kong, Suxian Zhao, Xingyu Li, and Guohong Fu. 2024.
\newblock Moddp: A multi-modal open-domain chinese dataset for dialogue discourse parsing.
\newblock In \emph{Findings of the Association for Computational Linguistics}, pages 10561--10573.

\bibitem[{Hao et~al.(2024)Hao, Wei, Cao, and Zhang}]{3.1}
Xiulan Hao, Shaohua Wei, Qian Cao, and Xiongtao Zhang. 2024.
\newblock Emotion recognition in conversations based on discourse parsing and graph attention network.
\newblock \emph{Telecommunications Science}, 40(5):100--111.

\bibitem[{Jiang et~al.(2023)Jiang, Li, He, Chen, Yu, and Zhang}]{DialogueDSA}
Yuru Jiang, Yu~Li, Weikai He, Jie Chen, Yanchao Yu, and Yangsen Zhang. 2023.
\newblock A new dataset and parsing model for chinese multiparty dialogue discourse structure.
\newblock In \emph{Proceedings of the 2023 International Conference on Asian Language Processing}, pages 221--227.

\bibitem[{Ju et~al.(2024)Ju, Zhang, Zhu, Li, Li, and Zhou}]{7}
Xincheng Ju, Dong Zhang, Suyang Zhu, Junhui Li, Shoushan Li, and Guodong Zhou. 2024.
\newblock Ecfcon: Emotion consequence forecasting in conversations.
\newblock In \emph{Proceedings of the 32nd ACM International Conference on Multimedia}, pages 2233--2241.

\bibitem[{Li et~al.(2024{\natexlab{a}})Li, Braud, Amblard, and Carenini}]{5.2}
Chuyuan Li, Chlo{\'e} Braud, Maxime Amblard, and Giuseppe Carenini. 2024{\natexlab{a}}.
\newblock Discourse relation prediction and discourse parsing in dialogues with minimal supervision.
\newblock In \emph{Proceedings of the 5th Workshop on Computational Approaches to Discourse}, pages 161--176.

\bibitem[{Li et~al.(2020)Li, Liu, Kan, Zheng, Wang, Lei, Liu, and Qin}]{molweni}
Jiaqi Li, Ming Liu, Min-Yen Kan, Zihao Zheng, Zekun Wang, Wenqiang Lei, Ting Liu, and Bing Qin. 2020.
\newblock Molweni: A challenge multiparty dialogues-based machine reading comprehension dataset with discourse structure.
\newblock In \emph{Proceedings of the 28th International Conference on Computational Linguistics}, pages 2642--2652.

\bibitem[{Li et~al.(2024{\natexlab{b}})Li, Song, Li, Zhang, and Hu}]{2.1}
Jingyang Li, Shengli Song, Yixin Li, Hanxiao Zhang, and Guangneng Hu. 2024{\natexlab{b}}.
\newblock Chatmdg: A discourse parsing graph fusion based approach for multi-party dialogue generation.
\newblock \emph{Information Fusion}, 110:102469.

\bibitem[{Li et~al.(2023)Li, Zhu, Shao, Yang, and Cambria}]{4}
Wei Li, Luyao Zhu, Wei Shao, Zonglin Yang, and Erik Cambria. 2023.
\newblock Task-aware self-supervised framework for dialogue discourse parsing.
\newblock In \emph{Findings of the Association for Computational Linguistics}, pages 14162--14173.

\bibitem[{Liu et~al.(2025)Liu, Li, Fan, and Zhu}]{DDPE}
Shannan Liu, Peifeng Li, Yaxin Fan, and Qiaoming Zhu. 2025.
\newblock Enhancing multi-party dialogue discourse parsing with explanation generation.
\newblock In \emph{Proceedings of the 31st International Conference on Computational Linguistics}, pages 1531--1544.

\bibitem[{Ma et~al.(2023)Ma, Zhang, and Zhao}]{4.1}
Xinbei Ma, Zhuosheng Zhang, and Hai Zhao. 2023.
\newblock Enhanced speaker-aware multi-party multi-turn dialogue comprehension.
\newblock \emph{IEEE/ACM Transactions on Audio, Speech, and Language Processing}, 31:2410--2423.

\bibitem[{Rennard et~al.(2024)Rennard, Shang, Vazirgiannis, and Hunter}]{1.2}
Virgile Rennard, Guokan Shang, Michalis Vazirgiannis, and Julie Hunter. 2024.
\newblock Leveraging discourse structure for extractive meeting summarization.
\newblock \emph{arXiv preprint arXiv:2405.11055}.

\bibitem[{Thompson et~al.(2024{\natexlab{a}})Thompson, Chaturvedi, Hunter, and Asher}]{LLaMIPa}
Kate Thompson, Akshay Chaturvedi, Julie Hunter, and Nicholas Asher. 2024{\natexlab{a}}.
\newblock Llamipa: an incremental discourse parser.
\newblock In \emph{Findings of the Association for Computational Linguistics}, pages 6418--6430.

\bibitem[{Thompson et~al.(2024{\natexlab{b}})Thompson, Hunter, and Asher}]{msdc}
Kate Thompson, Julie Hunter, and Nicholas Asher. 2024{\natexlab{b}}.
\newblock Discourse structure for the minecraft corpus.
\newblock In \emph{Proceedings of the 2024 Joint International Conference on Computational Linguistics, Language Resources and Evaluation}, pages 4957--4967.

\bibitem[{Wang et~al.(2024)Wang, Ji, and Kong}]{5.3}
Chengrui Wang, Shaoming Ji, and Fang Kong. 2024.
\newblock Local or global optimization for dialogue discourse parsing.
\newblock In \emph{CCF International Conference on Natural Language Processing and Chinese Computing}, pages 149--161. Springer.

\bibitem[{Xu et~al.(2024)Xu, Jiang, Gao, D'Haro, and Li}]{4.5}
Jiahui Xu, Feng Jiang, Anningzhe Gao, Luis~Fernando D'Haro, and Haizhou Li. 2024.
\newblock Unsupervised mutual learning of discourse parsing and topic segmentation in dialogue.
\newblock \emph{arXiv preprint arXiv:2405.19799}.

\bibitem[{Zhang et~al.(2023)Zhang, Chen, and Chen}]{3}
Duzhen Zhang, Feilong Chen, and Xiuyi Chen. 2023.
\newblock Dualgats: Dual graph attention networks for emotion recognition in conversations.
\newblock In \emph{Proceedings of the 61st Annual Meeting of the Association for Computational Linguistics}, pages 7395--7408.

\bibitem[{Zhang et~al.(2022)Zhang, Xu, Wang, Zhou, Zhao, and Teng}]{7.1}
Hanlei Zhang, Hua Xu, Xin Wang, Qianrui Zhou, Shaojie Zhao, and Jiayan Teng. 2022.
\newblock Mintrec: A new dataset for multimodal intent recognition.
\newblock In \emph{Proceedings of the 30th ACM International Conference on Multimedia}, pages 1688--1697.

\bibitem[{Zhao et~al.(2022)Zhao, Li, Wu, and He}]{JDDC}
Nan Zhao, Haoran Li, Youzheng Wu, and Xiaodong He. 2022.
\newblock Jddc 2.1: A multimodal chinese dialogue dataset with joint tasks of query rewriting, response generation, discourse parsing, and summarization.
\newblock In \emph{Proceedings of the 2022 Conference on Empirical Methods in Natural Language Processing}, pages 12037--12051.

\end{thebibliography}

\newpage 
\appendix

\begin{table*}
\centering
\setlength{\tabcolsep}{4pt} 
\resizebox{\textwidth}{!}{%
\begin{tabular}{ll@{\hspace{8.4pt}}c@{\hspace{5.6pt}}c@{\hspace{6pt}}c}
\toprule
\textbf{Discourse Relation} & \textbf{Definition} & \textbf{DraDDP (\%)} & \textbf{STAC (\%)} & \textbf{MODDP (\%)} \\
\midrule
Question-Answer Pair (QAP) & Arg2 is the answer to the question raised by Arg1. & 17.8 & 24.2 & 9.4 \\
Clarification\_Question (Clafi) & Arg2 provides clarification for Arg1. & 16.8 & 2.5 & 6.4 \\
Comment (Comt) & Arg2 expresses a viewpoint or evaluation of Arg1. & 14 & 17.6 & 16.4 \\
Acknowledgement (Ack) & Arg2 shows approval or acknowledgment for Arg1. & 9.6 & 9.6 & 2.7 \\
Contrast (Contrast) & Arg1 and Arg2 differ on a shared theme. & 7.6 & 4.7 & 6.6 \\
Elaboration (Elab) & Arg2 elaborates on Arg1 in detail. & 6.8 & 8.3 & 15.7 \\
Correction (Corr) & Arg2 corrects Arg1. & 5.9 & 2 & 0.2 \\
Continuation (Continu) & Arg2 is a continuation of the content of Arg1. & 4.8 & 9.4 & 9.3 \\
Explanation (Expl) & Arg2 provides an explanation for Arg1. & 3.5 & 4.2 & 5.2 \\
Narration (Narr) & Arg2 narrates or describes Arg1. & 3.5 & 1.2 & 1.8 \\
Parallel (Paral) & Arg1/2 have alike semantic structures and themes. & 2.2 & 2 & 2.3 \\
Q-Elab (Q-Elab) & Arg1 is a question, and Arg2 elaborates on it in detail. & 2.1 & 5.7 & 6.3 \\
Result (Resl) & Arg2 is the result of the situation described in Arg1. & 2.1 & 5.5 & 6.8 \\
Conditional (Condi) & Arg2 is the condition for Arg1. & 1.6 & 1.2 & 3.2 \\
Alternation (Alte) & Arg1 and Arg2 represent interchangeable situations. & 1 & 1.4 & 4.5 \\
Background (Backg) & Arg2 provides background information for Arg1. & 0.7 & 0.6 & 3.2 \\
\bottomrule
\end{tabular}%
}
\caption{Definition and proportion of discourse relations in DraDDP, MODDP, and STAC datasets.}
\label{tab:discourse_relations}
\end{table*}

\section{Discourse Relations}
\label{sec:appendix1}
Table~\ref{tab:discourse_relations} presents the 16 discourse relations used in our annotation. For each relation, we provide its definition and distribution across the DraDDP, STAC, and MODDP datasets. The percentages in Table~\ref{tab:discourse_relations} reveal significant variations in how frequently different relations appear, reflecting the unique conversational focuses of each dataset. Notably, \emph{Question-Answer Pair} relation dominates in DraDDP and STAC (17.8\% and 24.2\%, respectively), but is far less common in MODDP (9.4\%), suggesting the latter involves fewer direct Q\&A exchanges. Conversely, MODDP shows a higher prevalence of the \emph{Elaboration} (15.7\%) and \emph{Continuation} (9.3\%) relations, indicating a stronger tendency toward extended, descriptive turns. \emph{Acknowledgment (Ack)} is prominent in both DraDDP and STAC (9.6\% each) but minimal in MODDP (2.7\%), potentially pointing to differences in interactive feedback or social rapport. Furthermore, the high frequency of \emph{Clarification\_Question} in DraDDP (16.8\%) compared to the other two datasets highlights its distinctive focus on resolving misunderstandings.

\section{Analysis on Controversial Cases}
\label{sec:appendix2}
To illustrate the complexity of discourse relation annotation and the challenges faced by annotators, we selected several controversial cases from the annotation process for analysis. These cases demonstrate the subtle boundaries between different discourse relation types and highlight the difficulty of achieving perfect agreement among annotators.

\begin{table}[htbp]
\centering
\resizebox{\linewidth}{!}{%
\begin{tabular}{lccccc}
\toprule
\textbf{Speaker Count} & \textbf{D=1} & \textbf{D=2} & \textbf{D=3} & \textbf{D=4} & \textbf{D$\geq$5} \\
\midrule
$s\leq2$ & 88.6\% & 9.0\% & 1.2\% & 0.8\% & 0.3\% \\
$2<s\leq6$ & 80.2\% & 13.5\% & 3.0\% & 1.5\% & 1.8\% \\
$s>6$ & 78.1\% & 12.6\% & 3.3\% & 2.1\% & 3.9\% \\
\bottomrule
\end{tabular}%
}
\caption{Distribution of dependency relation distances across different speaker counts.}
\label{tab:dependency_speaker}
\end{table}

\begin{figure*}
    \centering
    \includegraphics[width=0.9\linewidth]{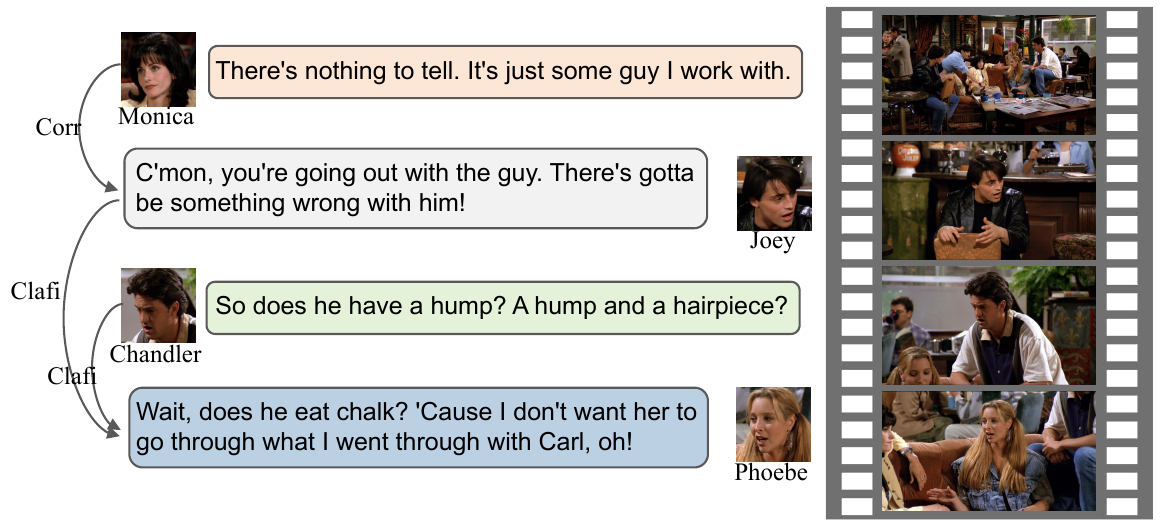} 
    \caption{Controversial annotation cases in the DraDDP dataset.}
    \label{fig:Controversial Annotation Cases}
\end{figure*}

As shown in Figure~\ref{fig:Controversial Annotation Cases}, although annotators reached consensus on the dependency structures between utterances, they disagreed on the specific classification of relation types. Regarding the relation between utterances $u_{1}$ and $u_{0}$, some annotators classified it as \emph{Comment}, believing that $u_{1}$ expressed Joey's evaluative opinion about Monica's situation. Other annotators tended to mark it as \emph{Contrast}, considering the contrasting relation between romantic partners and colleagues. After thorough discussion, we ultimately classified this relation as \emph{Correction}, because $u_{1}$ actually corrects Monica's description of her relationship status, correcting the nature of her relationship with colleagues to a romantic one. For the relation between utterances $u_{2}$ and $u_{1}$, some annotators considered this to be \emph{Elaboration}, interpreting Chandler's question as providing specific details about the ``certain mistakes'' mentioned in $u_{1}$. Other annotators argued this was a \emph{Clarification\_Question}, believing that $u_{2}$ seeks to clarify the specific meaning of the ``certain mistakes'' Joey mentioned. We ultimately chose the \emph{Clarification\_Question} label, because the primary purpose of $u_{2}$ is to clarify the nature of the problem Joey implied, similar to $u_{3}$.

The above examples show that although each type of discourse relation has clear definitional distinctions, in real daily conversations, they often lack clear boundaries and have semantic overlaps and ambiguous areas. This phenomenon illustrates the difficulty of the annotation task and the complexity of multimodal dialogue discourse structure parsing. In Table~\ref{tab:Distinction between Discourse Relations}, we have identified five groups of discourse relations that are easily confused during the annotation process and clarified their core criteria for discrimination, aiming to provide a reference for subsequent work.

\section{Impact of Participant Numbers on Dependency Distance}
\label{sec:appendix3}
To comprehensively validate the relation between participant numbers and dialogue complexity, we analyzed the correlation between dependency distance and speaker count. As shown in Table~\ref{tab:dependency_speaker}, we divided the dialogues in the DraDDP dataset into three groups based on the number of speakers ($s\leq2$, $2<s\leq6$, $s>6$) and calculated the distribution of different dependency distances ($d$) within each group.

With the increase in speaker count, the proportion of local dependencies gradually decreases, while the proportion of long-distance dependency increases significantly. This trend fully validates the characteristic that topic branching and switching occur more frequently in multi-party dialogues. This means that in scenarios with more participants, models need to traverse more dialogue history turns to capture authentic discourse dependency structures, thereby significantly increasing the difficulty of discourse relation identification and parsing.

\section{Baselines}
\label{sec:appendix3.1}
We employed four advanced dialogue discourse parsing systems (RLTST \cite{4.2}, BERTLine \cite{4.7}, MODDP \cite{MODDP}, and LLaMIPa \cite{LLaMIPa}) to validate the effectiveness and applicability of the DraDDP dataset.

RLTST~\cite{4.2}: A multi-task learning-based dialogue discourse parser that integrates complementary information from dialogue discourse parsing and addressee recognition, alleviating data sparsity issues without requiring additional manual annotations. Its core mechanism employs reinforcement learning to filter samples with significant gains in addressee recognition, and utilizes a task-aware structure transformer to distinguish between task-shared and task-private structures, avoiding mutual interference.

BERTLine~\cite{4.7}: It fine-tunes BERT to directly encode EDU pairs and employs simple linear layers to perform dependency structure predictions. This model is capable of handling multi-parent structures and can effectively capture complex dependencies found in multi-turn dialogues.

\begin{table}[htbp]
\centering
 \setlength{\tabcolsep}{16pt}
\begin{tabular}{lcc}
\toprule
Modalities & Link & Link\&Rel \\
\midrule
T & 84.67 & 53.69 \\
V & 43.38 & 22.21 \\
A & 47.39 & 38.83 \\
T+V & 83.61 & 52.97 \\
T+A & 84.83 & 54.76 \\
V+A & 50.12 & 40.39 \\
T+A+V & 84.55 & 53.34 \\
\bottomrule
\end{tabular}
\caption{Ablation results (F1-score) of different modalities.}
\label{tab:multimodal_results}
\end{table}

MODDP~\cite{MODDP}: It is a discourse parser for multimodal open-domain Chinese dialogues, which employs RoBERTa, ViT, and Wav2Vec2.0 to encode textual, visual, and audio modality features respectively, and achieved multimodal information interaction through cross-modal multi-head attention mechanisms. This model introduces context-aware and speaker-aware dialogue interaction modules to enhance overall understanding of dialogue structure.

LLaMIPa \cite{LLaMIPa}: It is an incremental parser based on fine-tuned LLaMA3. When processing each new utterance unit, the model not only considers the current textual content but also incorporates previously predicted discourse structure information to synchronously perform dependency structure and relation prediction, representing a strong baseline model in the dialogue discourse parsing field.

\section{Ablation Study}
\label{sec:appendix5.1}
To explore the contributions of different modalities in the DraDDP dataset, we conducted ablation experiments based on the LLaMIPa$^\ddagger$ (Qwen2.5-Omni) model. As shown in Table~\ref{tab:multimodal_results}, under the unimodal setting, the text modality performs best (Link-F1 of 84.67 and Link\&Rel-F1 of 53.69), far surpassing the visual and audio modalities. This indicates that text carries the most direct and complete semantic information and serves as the foundation for identifying discourse relations.

Among bimodal combinations, text+audio achieves optimal performance on the test set, with Link\&Rel-F1 reaching 54.76, an improvement of 1.07 over pure text, significantly outperforming the text+visual combination. This is mainly because the audio modality more directly conveys speakers' intonation changes, emotional tendencies, and voice intensity, which have strong correlations with semantic relations between utterances (e.g., \emph{Acknowledgement}, \emph{Comment}, \emph{Clarification\_Question}, etc.) (see \cref{subsec:Error Pattern Analysis} and \cref{subsec:Case Study}). In contrast, although the visual modality provides cues such as facial expressions and body movements, its correlation with semantic relations is relatively weak, and complex scene backgrounds introduce substantial visual noise, which to some extent interferes with the model's semantic relation judgment.

When fusing all three modalities, model performance is slightly lower than the pure text modality, indicating that under the current model architecture and video processing approach, the introduction of visual information partially offsets the performance gains brought by audio information. As analyzed in \cref{subsec:Experimental Results}, the effectiveness of multimodal information exhibits significant differences across various dialogue scenarios. In two-party dialogues ($s\leq2$), visual cues such as facial expressions and eye contact are more focused and prominent, enabling the video modality to effectively enhance relation type identification. As the number of participants increases ($s>6$), visual information becomes more scattered and noisier, while audio signals (e.g., speakers' intonation variations and emotional intensity) demonstrate stronger advantages in capturing discourse dependencies and semantic relations in complex multi-party interactions. Therefore, how to dynamically weight different modalities according to dialogue contextual features and more effectively extract spatiotemporal video features will be an important direction for future multimodal dialogue discourse parsing research.

\section{Evaluation on Different LLMs}
\label{sec:appendix5}
To further evaluate the performance of the DraDDP dataset on current mainstream LLMs, we selected Claude-Sonnet-4\footnote{\url{https://www.anthropic.com/claude/sonnet}}, GPT-4o\footnote{\url{https://openai.com/}}, and Qwen3-235B\footnote{\url{https://huggingface.co/Qwen/Qwen3-235B-A22B-Instruct-2507}} as representative unimodal text large models, while selecting Gemini 2.5 Pro\footnote{\url{https://gemini.google.com/}} and Qwen3-VL-235B\footnote{\url{https://huggingface.co/Qwen/Qwen3-VL-235B-A22B-Instruct}} as representative multimodal large models. For model input, we adopted a prompt template consistent with the ``Vanilla'' type in \citet{5.1}, which only provides textual dialogue content. For multimodal models, we additionally provided audio-visual content and specified in the template: ``\emph{Analyze the following dialogue by integrating textual content with multimodal cues from the video, including but not limited to speakers' facial expressions, body language, tone variations, and emotional signals. Identify discourse dependency structures and relation types, then complete the annotation according to the specified format.}''.

\begin{table}[t]
  \centering
  \small
  \resizebox{\linewidth}{!}{
  \begin{tabular}{l l c c}
    \toprule
  {Category} & {Model} & 
     Link & Link\&Rel \\
    \midrule
    \multirow{3}{*}{Unimodal} & Claude-Sonnet-4 & 71.46 & 23.18 \\
    & GPT-4o & 78.11 & 35.87 \\
    & Qwen3-235B & 71.48 & 42.18 \\
    \hline
    \multirow{2}{*}{Multimodal} & Gemini 2.5 Pro & 68.17 & 21.69 \\
    & Qwen3-VL-235B & 71.64 & 37.71 \\
    \bottomrule
  \end{tabular}
  }
  \caption{Baseline performance of LLM on the DraDDP dataset (F1 score).}
  \label{tab:model_data}
\end{table}

The experimental results are shown in Table~\ref{tab:model_data}. We observe that even on the most advanced large models currently available, the parsing performance on DraDDP remains significantly lower than specialized models, particularly showing poor performance in Link\&Rel-F1. This finding echoes the research conclusions of \citet{5} and \citet{5.1}, further validating the extremely high complexity of multi-party dialogue discourse parsing.

In the pure text modality, GPT-4o performs best on linking (Link at 78.11\%), while Qwen3-235B performs relatively better on relation (Link\&Rel-F1 at 42.18\%), reflecting the differences in models' capabilities to capture dialogue structure and semantic relations.

Notably, the multimodal models Gemini 2.5 Pro and Qwen3-VL-235B did not demonstrate significant advantages over the pure text models on this task. This phenomenon may stem from the following reasons: 1) general-purpose multimodal large models lack targeted training on the specific task of dialogue discourse parsing; 2) the models may introduce semantic interference when fusing text-video cross-modal information; 3) current large models' capabilities in deep fusion of multimodal information and understanding of complex dialogue structures remain immature.

\begin{table*}[ht]
\centering
\begin{tabular}{p{1.3cm} p{7.2cm} p{6.6cm}}
\hline
\textbf{Type} & \textbf{Distinction} & \textbf{Example} \\
\hline
\emph{Clafi} & \emph{Clarification\_Question} is to ask about the ``current situation, details, or ambiguities of an existing statement'' to confirm facts or eliminate information bias. & $u_{0}$: There's nothing to tell. It's just some guy I work with. \newline $u_{1}$: C'mon, you're going out with the guy. There's gotta be something wrong with him! \\
\emph{Q-Elab} & \emph{Q-Elab} is to further inquire or expand on an ``already raised question'' to obtain more comprehensive information, focusing on the ``question itself''. & $u_{0}$: Who's Paul? \newline $u_{1}$: Paul, the wine guy? Paul? \\
\hline
\emph{Continu} & \emph{Continuation} is generally a ``parallel extension'' to the current topic, with no strict order, which is just horizontal expansion of the topic. & $u_{0}$: I say push her down the stairs. \newline $u_{1}$: Push her down the stairs! Push her down the stairs! Push her down the stairs! \\
\emph{Narr} & \emph{Narration} is a ``vertical expansion'' of the current topic, with a clear implicit sequence, process, or logical connection, presenting complete events or information. & $u_{0}$: Yeah, right! See, he gave up something, but then he got those magic beans. \newline $u_{1}$:  And then he woke up, and there was a big plant outside his window, full of possibilities and stuff... \\
\hline
\emph{Expl} & \emph{Explanation} provides a clear reason, basis, or reasoning process for a ``certain viewpoint, phenomenon, or result'', focusing on ``answering why''. & $u_{0}$: Oh my god, oh, you guys are great. \newline $u_{1}$: We all chipped in.\\
\emph{Backg} & \emph{Background} has a weaker causal relationship, providing relevant background information (such as time, scene, premise, etc.) for the ``current topic''. & $u_{0}$: I'm smoking. I'm smoking, I'm smoking. \newline $u_{1}$: Oh, I can't believe you! You've been so good, for three years! \\
\hline
\emph{Corr} & \emph{Correction} directly points out and gives the correct content for ``errors in the other party's statement'', focusing on ``correcting errors'', can be replaced with ``no no no''. & $u_{0}$: ...That's it. I'm getting cigarettes. \newline $u_{1}$:  No no no! \\
\emph{Contrast} & \emph{Contrast} highlights the difference or opposition, focusing on ``presenting differences'', not involving ``error correction''. & $u_{0}$: How does she do that? \newline $u_{1}$: I cannot sleep in a public place, libraries, airplanes, movie theaters.... \\
\hline
\emph{Ack} & The two types have a ``containment and independence'' relationship. \emph{Acknowledgement} expresses clear agreement with the other's statement/viewpoint, and is very brief. & $u_{0}$: Would you look at her? She is so peaceful. \newline $u_{1}$: yeah. \\
\emph{QAP} & \emph{Question-Answer Pair} may include a brief acknowledgement and then add specific information. In such cases, ``yes'' serves merely as a prefix to the answer, and the whole is still classified as \emph{Question-Answer Pair}. & $u_{0}$: Alright. Phoebe? \newline $u_{1}$: Okay, okay. If I were omnipotent for a day, I would want, um, world peace, no more hunger, good things for the rain forest. And bigger boobs. \\
\hline
\end{tabular}
\caption{Distinction between easily confusable discourse relations.}
\label{tab:Distinction between Discourse Relations}
\end{table*}

\end{document}